\pdfoutput=1

\documentclass[letterpaper, 10 pt, conference]{ieeeconf}  

\IEEEoverridecommandlockouts                              

\usepackage{graphics} 
\usepackage{epsfig} 
\usepackage{mathptmx} 
\usepackage{times} 
\usepackage{amsmath} 
\usepackage{amssymb}  
\usepackage{color}
\usepackage{todonotes}
\usepackage{algorithm}
\usepackage{algpseudocode}
\usepackage{caption}
\usepackage{afterpage}
\usepackage{placeins}
\usepackage{booktabs}
\usepackage{siunitx}
\usepackage{float}
\usepackage{flushend}
\usepackage{algorithmicx}
\usepackage{algorithm}
\usepackage{algpseudocode}
\usepackage{hyperref}

\usepackage[labelfont=bf]{caption}

\newcommand{\etal}{\textit{et al}. }
\newcommand{\ie}{\textit{i}.\textit{e}., }
\newcommand{\eg}{\textit{e}.\textit{g}. }

\algnewcommand\algorithmicinput{\textbf{Input:}}
\algnewcommand\Input{\item[\algorithmicinput]}
\algnewcommand\algorithmicoutput{\textbf{Output:}}
\algnewcommand\Output{\item[\algorithmicoutput]}

\usepackage{forloop}
\newcounter{ct}

\newcommand{\doubleQuote}[1]{\lq\lq{#1}\rq\rq}

\newcommand{\w}{\mathbf{w}}
\newcommand{\T}{\mathsf{T}}

\graphicspath{{figures/}}

\graphicspath{{figures/}}

\title{\LARGE \bf
Sports Camera Calibration via Synthetic Data
}

\author{Jianhui Chen and  James J. Little 
\thanks{Jianhui Chen and James J. Little are with the Department of Computer Science, University of British Columbia, Vancouver, Canada. {\tt\small \{jhchen14,little\}@cs.ubc.ca}}
}        

\begin{document}

\maketitle

\thispagestyle{empty}
\pagestyle{empty}

\begin{abstract}
Calibrating sports cameras is important for autonomous broadcasting and sports analysis. Here we propose a highly automatic method for calibrating sports cameras from a single image using synthetic data. First, we develop a novel camera pose engine. The camera pose engine has only three significant free parameters so that it can effectively generate a lot of camera poses and corresponding edge (\ie field marking) images. Then, we learn compact deep features via a siamese network from paired edge image and camera pose and build a feature-pose database. After that, we use a novel two-GAN (generative adversarial network) model to detect field markings in real images. Finally, we query an initial camera pose from the feature-pose database and refine camera poses using truncated distance images. We evaluate our method on both synthetic and real data. Our method not only demonstrates the robustness on the synthetic data but also achieves the state-of-the-art accuracy on a standard soccer dataset and very high performance on a volleyball dataset. 
\end{abstract}


\section{Introduction}
Camera calibration is a fundamental task for robotic applications such as visual SLAM and relocalization. It also provides training examples for data-driven applications such as autonomous broadcasting systems which replace human operators with robotic cameras to capture sports events. For example, companies like ESPN \cite{chen2017where} train intelligent broadcast systems using accurate camera poses. Our task is to estimate the camera poses from a single color image in sports such as soccer games.     

Many methods were developed to align images with templates (\eg soccer pitch). The current trend is using highly/fully automatic methods to deal with large-scale data. For example, Wen \etal \cite{wen2016court} first reconstructed a panoramic court image from a basketball video. Then, they warped the panoramic court to the court template. However, their method requires multiple images from well-textured stadiums (\eg for basketball games). 

Our work is closely related to \cite{homayounfar2017sports} and \cite{sharma2018automated} but we approach the problem from a very different perspective. The deep structured model (DSM) \cite{homayounfar2017sports} uses semantic information such as grassland and field markings to relocalize the soccer field. On the other hand, Sharma \etal \cite{sharma2018automated} uses a synthetic dictionary to query the camera pose. Both of them use the standard projection (2D to 2D homography or a standard pinhole camera model) in their methods. None of them approaches the problem by analyzing the specific form of sports camera poses. We will show that with a better decomposition of sports camera poses, we can obtain more accurate calibration results.

\begin{figure}[t]
	\centering
	\includegraphics[width=1.0\linewidth]{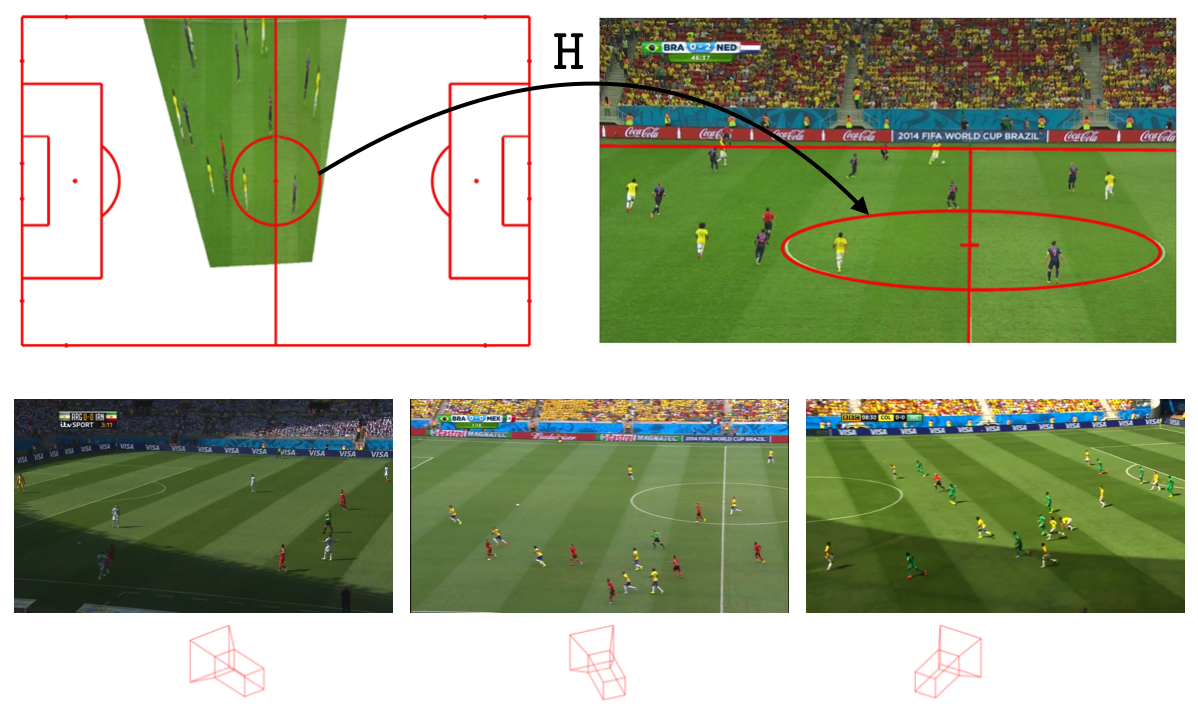} 
	\caption{Sports camera calibration and camera location prior. The first row shows our task of calibrating a sports camera from a single image. The second row shows our observation that these cameras have a strong prior of the location (roughly shown by red cameras). Our work develops a novel camera calibration method by using the location prior.}    
	\label{fig:motivation}
    \vspace{-0.15in}
\end{figure}

\begin{figure*}[t]
	\centering
	\includegraphics[width=0.95\linewidth]{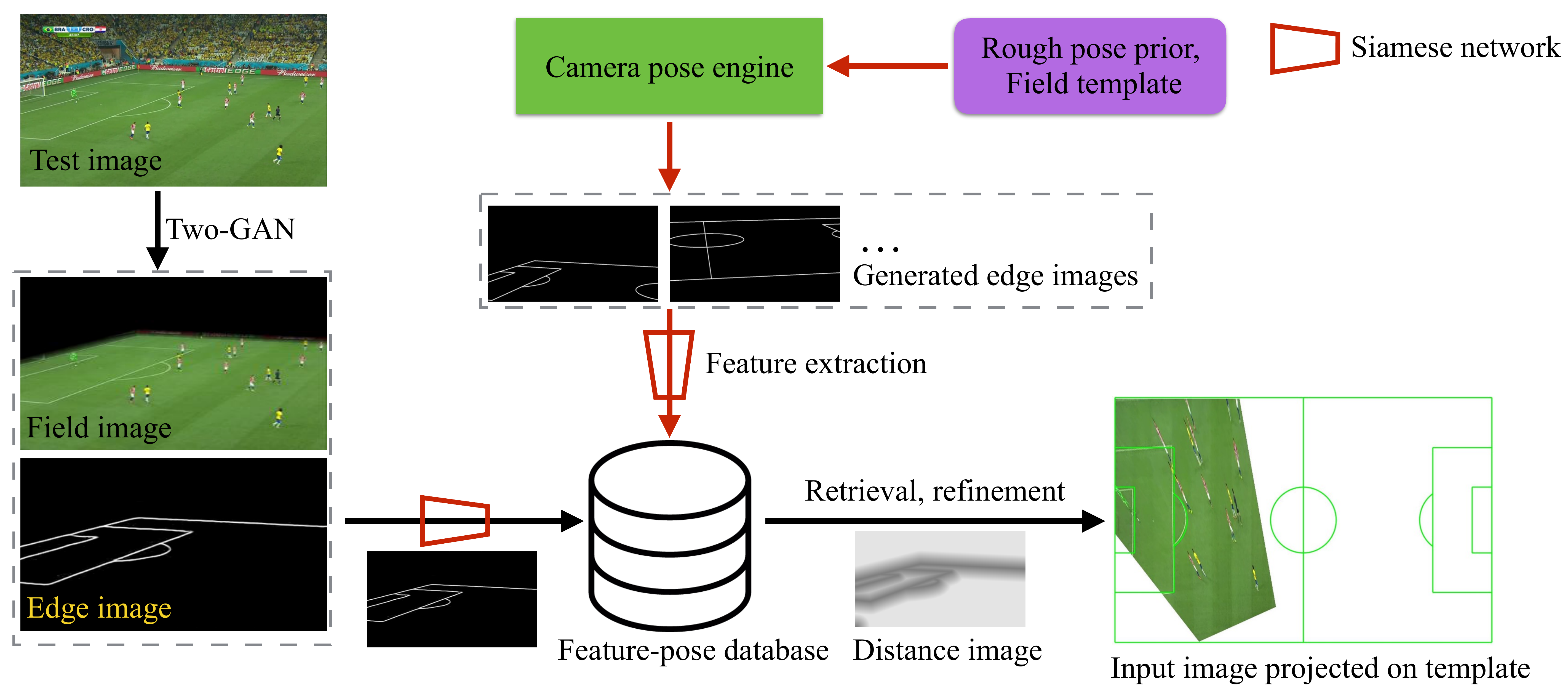}
	\caption{Sports camera calibration pipeline. In training (red arrows), we use a novel camera pose engine to synthesize a set of edge images which are embedded into a low dimensional feature space. In this way, we obtain a feature-pose database. In testing (black arrows), we detect field markings using a two-GAN model from a testing image. The camera pose is quickly retrieved from the database and is refined using the distance image. Best viewed in color.}
    \vspace{-0.15in}
	\label{fig:pipeline}
\end{figure*}

Our observation of sports images inspires our method. Figure \ref{fig:motivation} (second row) shows image examples from the WorldCup dataset \cite{homayounfar2017sports}. Our first observation is that field markings (\eg lines and circles) must be used in the calibration as they are the visible evidence of camera poses. The second observation is that the camera poses have some \doubleQuote{default} settings. For example, the cameras are roughly located near and above the middle line. The cameras have a large range of pan angles (from left to right) and a small range of tilt angle (from top to down). By watching lots of broadcasting videos, we found these \doubleQuote{default} settings hold for many sports videos. We will show how to use these \doubleQuote{default} settings as a prior to simplify the camera model.

Our method extends the two-point method \cite{chen2018two} by relaxing the constraint of the camera location. The two-point method assumes the camera location is \textit{exactly} known in training/testing. However, this assumption generally does not hold when cameras change positions in different games. In this work, we only assume the camera location is \textit{roughly} known. As a result, our method can be applied to more situations than the two-point method. For instance, the images in training and testing can be from different stadiums. 

Figure \ref{fig:pipeline} shows the pipeline of our method. We first build a feature-pose database in training. In the database, the camera poses are generated by a novel camera pose engine. The features are learned from a siamese network. In testing, our method detects field markings from the input image. Then, an initial camera pose is retrieved from the database and is refined using distance images. In summary, our paper has three contributions:
\begin{itemize}
\setlength{\itemsep}{2pt}	
  	\item We propose a novel sports camera pose engine that only has three significant free parameters.
  	\item We also propose an effective feature extraction method for edge images and an end-to-end two-GAN model to detect field markings.   
    \item We demonstrate the state-of-the-art performance on a standard soccer dataset and very high performance on a volleyball dataset. 
\end{itemize}

In the following, we start with a discussion of related work and then describe our method. Finally, we demonstrate our method on both synthetic and real datasets.

\section{Related Work}
{\bf Camera Relocalization and Sports Camera Calibration:}
Camera relocalization has been widely studied in the context of global localization for robots using edge images \cite{qiu2017model}, random forests \cite{meng2017backtracking,cavallari2017onthefly} and deep networks \cite{kendall2015posenet,rubio2015efficient,naseer2017deep,valada2018deep}. 

In sports camera calibration, researchers assume the playing surface is flat so that camera calibration is equivalent to estimating the homography from the ground to the image. Previous work \cite{puwein2011robust,gupta2011using,ghanem2012robust} first manually annotates several reference images. Then, they calibrate other images by finding correspondences from reference images. Fully-automatic methods are emerging because they require no or fewer user interactions \cite{wen2016court,zeng2018calibrating}. For example, Homayounfar \etal \cite{homayounfar2017sports} formulate the problem as a branch and bound inference in a Markov random field where an energy function is defined in terms of semantic cues (\eg field surface, lines and circles). 

Recently, Sharma \etal \cite{sharma2018automated} formulate the problem as a nearest neighbour search problem over a synthetic dictionary of edge images. The method first warps the training image to the field template in which the training image becomes a convex quadrilateral. Then, it simulates pan, tilt and zoom by manipulating the quadrilateral. This approach also extensively studies different representations for edge images and finally chooses the histogram of gradients (HOG) features. Our method is significantly different from this method in two points. First, we propose a novel camera pose engine to generate edge images. The camera pose engine is physically interpretable and only requires very rough information of camera locations. Second, we learn features of edge images using a siamese network that provides much more compact features than HOG. 

{\bf Edge Image Detection and Representation:}
An edge image is the projection of a template (field markings) in an image. Previous work developed color-based kernel \cite{hayet2004robust}, line and ellipse detection \cite{puatruaucean2012parameterless} to distinguishes field-marking pixels from other pixels. Recently, Sharma \etal \cite{sharma2018automated} uses a conditional generative adversarial network (CGAN)~\cite{isola2017image} to directly generate edge images from an RGB image. 

Edge images are represented by features that are much more efficient than raw edge images in search \cite{thomas2007real,homayounfar2017sports}. For example, Sharma \etal \cite{sharma2018automated} used Chamfer transformation and HOG to represent edge images for soccer games. In an early version of \cite{sharma2018automated}, deep features have been reported with good results on synthetic experiments but poor results on real data experiments. 

Metric learning was used to learn deep features from paired/triplet images \cite{hadsell2006dimensionality,loquercio2017efficient}. For example, Wohlhart and Lepetit \cite{wohlhart2015learning} learns deep features from a triplet network to estimate 3D poses for small indoor objects. Doumanoglou \etal \cite{doumanoglou2016siamese} analyze the influence of different objectives such as regression loss of poses and contrastive loss of paired images. To the best of our knowledge, few previous methods focus on edge images which are texture-less and much more challenging than well-textured images.

\section{Method}
Our method has a camera pose engine, a deep feature extractor, a two-GAN model for field marking detection and a camera pose refinement process. Limited by pages, intermediate results from each step are provided in the supplementary video. 

\subsection{Camera Pose Engine}
\label{sec:camera_model}
We use the pinhole camera model to describe the projective aspects of a camera
\begin{equation}
	\mathtt{P} = \mathtt{K} \mathtt{R} [\mathtt{I} | -\mathbf{C}],
	\label{equ:pinhole}
\end{equation}
\noindent where $\mathtt{K}$ is the intrinsic matrix, $\mathtt{R}$ is a rotation matrix from world to camera coordinates, $\mathtt{I}$ is an identity matrix and $\mathbf{C}$ is the camera's center of projection in the world coordinate. To simplify the problem, we assume square pixels, a principal point at the image center and no lens distortion. We found these assumptions hold well for our problem. As a result, the focal length $f$ is the only unknown variable in the intrinsic matrix.

Most sports cameras are pan-tilt-zoom (PTZ) cameras \cite{thomas2007real,chen2018two}. As a result, we decompose the rotation matrix $\mathtt{R}$ in \eqref{equ:pinhole}:
\begin{equation}
	\mathtt{P} = 
	\underbrace{\mathtt{K} \mathtt{Q}_{\phi}\mathtt{Q}_{\theta}}_\text{PTZ} 
	\underbrace{\mathtt{S} [\mathtt{I} | -\mathbf{C}]}_\text{prior}.
	\label{equ:ptz}
\end{equation}
\noindent The combination of $\mathtt{Q}_{\phi}\mathtt{Q}_{\theta} \mathtt{S}$ describes rotations from world to camera coordinates. First, it rotates the camera to the PTZ camera base (\eg a big tripod) by $\mathtt{S}$. Then the camera pans by $\mathtt{Q}_{\theta}$ and tilts by $\mathtt{Q}_{\phi}$. $\mathtt{P}$ can be separated into two parts. The right part $\mathtt{S} [\mathtt{I} | -\mathbf{C}]$ is time invariant because camera bases are generally fixed to capture stable videos in sports.

To reduce the number of free parameters, we further decompose the base rotation $\mathtt{S}$ to
\begin{equation}
\label{equ:base_rotation}
\mathtt{S} = \mathtt{S}_{\theta^\prime} \mathtt{S}_\rho \mathtt{S}_{\phi^\prime},
\end{equation}
where $\phi^\prime$, $\rho$, and $\theta^\prime$ are tilt, roll and pan angles of the camera base. We simplify the formation by eliminating $\mathtt{S}_{\theta^\prime}$ since $\mathtt{Q}_{\theta}\mathtt{S}_{\theta^\prime}$ can be represented by $\mathtt{Q}_{\theta}$ with $\theta = \theta + \theta^\prime$. Thus, the camera model becomes:
\begin{equation}
	\mathtt{P} = 
	\underbrace{\mathtt{K} \mathtt{Q}_{\phi}\mathtt{Q}_{\theta}}_\text{PTZ} 
	\underbrace{\mathtt{S}_\rho \mathtt{S}_{\phi^\prime} [\mathtt{I} | -\mathbf{C}]}_\text{prior}.
	\label{equ:pinhole_base_rotation}
\end{equation}
\begin{figure}[t]
	\centering
	\includegraphics[width=0.9\linewidth]{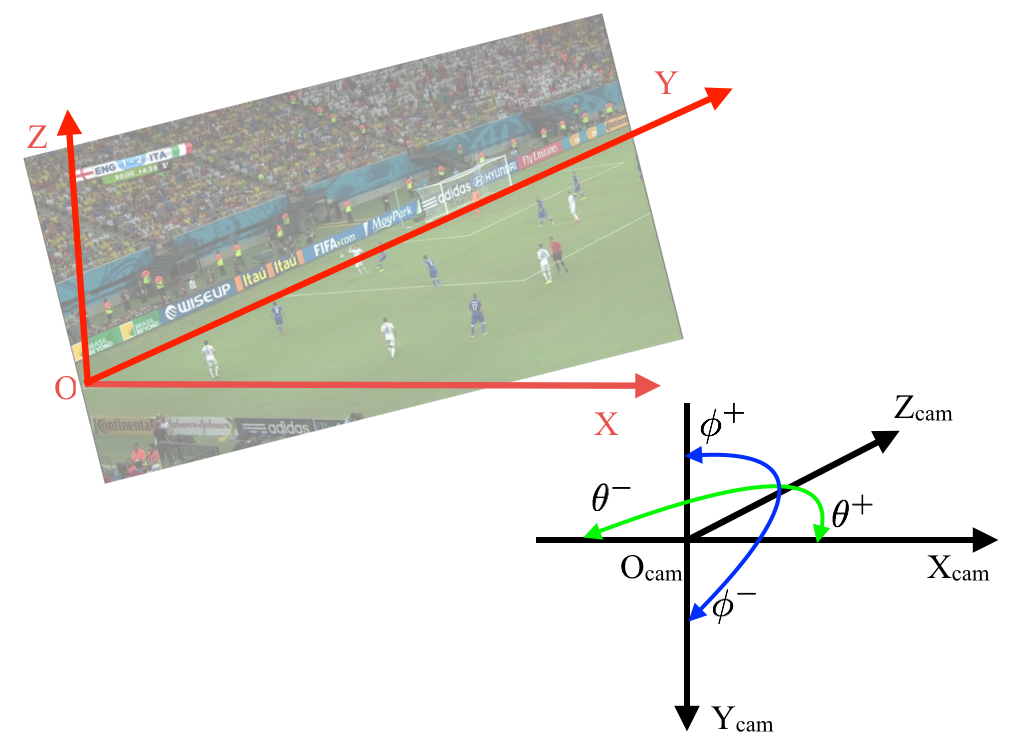}    
	\caption{World (red) and camera (black) coordinates used in \eqref{equ:pinhole_base_rotation}. The origin of the world coordinate is at the left bottom of the soccer template. The Z-axis in the camera coordinate is aligned with the Y-axis in the world coordinate. Best viewed in color.}    
	\label{fig:coord}
    \vspace{-0.15in}
\end{figure}
\noindent Without loss of generality, we set the world and camera coordinates as in Figure \ref{fig:coord}. The world origin is at the left bottom of the field template. When pan and tilt are zeros, the camera looks along the Y-axis of the world coordinate.

At first glance, \eqref{equ:pinhole_base_rotation} seems to have many free parameters (3 for location, 4 for rotation and 1 for focal length). However, the number of significant free parameters is only three. In \eqref{equ:pinhole_base_rotation}, we set $\phi^\prime = -90^{\circ}$ with which cameras are set up to be \doubleQuote{level} because the effect of $\mathtt{S}_{\phi^\prime}$ will be canceled by $\mathtt{Q}_{\phi}$. $\rho^\prime$ varies in a small range ($\pm 0.1^{\circ}$) because the camera base prevents the camera from rotating about its direction-of-view. For sports fields, $\mathbf{C}$ is further constrained in practice, for example, most cameras are above and along the center line for soccer games. As a result, the significant free parameters are $f$, $\phi$, $\theta$ whose ranges are known (\eg from training data). We generate lots of camera poses and paired edge images by uniformly sampling these free parameters. 

\subsection{Deep Edge Feature Extraction}
\label{sub_sec:deep_feature}
Given edge images and their poses, we learn compact features via a siamese network \cite{hadsell2006dimensionality}. The input of the network is a pair of edge images. The label is similar or dissimilar. A pair of edge images is set as similar if their pan, tilt and focal length differences are within pre-defined thresholds, and vice versa. The siamese network has two branches each of which is a convolutional neural network $f_\w(\cdot)$. We want to ensure an image $x_1$ is closer to another image $x_2$ when the camera poses are similar. The loss function is:
\begin{equation}
\begin{split}
\mathcal{L}(\w, x_1, x_2, y) &= yD_{\w}(x_1, x_2) \\
                              &+ (1-y)\max(0, m - D_{\w}(x_1, x_2)),
\end{split}
\end{equation}
where $D_{\w}(x_1, x_2) = \|f_\w(x_1) - f_\w(x_2) \|_2^2$, y is the similar/dissimilar label and $m$ is a margin (1 in this work). 

We have tried many network structures for the siamese network. We found max-pooling and batch normalization have slightly negative impacts on the performance, which agrees with \cite{doumanoglou2016siamese}. The network consists of 5 stride-2 convolutions (kernel size 7, 5, 3, 3, 3) followed by a $6 \times 10$ convolution and a $L2$ normalization layer. The learned feature dimension is experimentally set as 16. The siamese network also extracts features from detected edge images in the testing.

\subsection{Two-GAN Model for Field Marking Detection}
\begin{figure*}[t]
	\centering
	\includegraphics[width=0.95\linewidth]{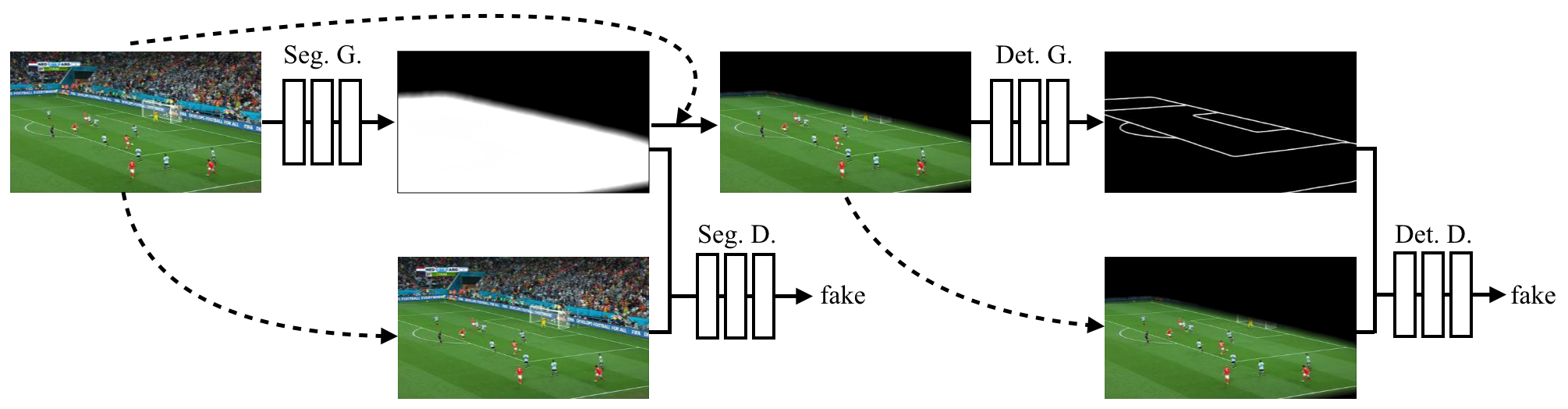}    
	\caption{The two-GAN model. We use a two chained conditional GAN to detect field markings. \doubleQuote{Seg.} and \doubleQuote{Det.} are short for segmentation and detection, respectively. \doubleQuote{G.} and \doubleQuote{D.} are short for generative network and discriminative network, respectively. }    
	\label{fig:two_cgan_pipeline}
    \vspace{-0.15in}
\end{figure*}

We extend the work of \cite{sharma2018automated} to detect field markings using a two-GAN model. We first use a segmentation GAN to segment the playing surface from the whole image. Then, we use a detection GAN to detect field markings from the playing surface. The motivation of using two GANs is to avoid the negative influence of background objects (\eg white lines on commercial boards).

Figure \ref{fig:two_cgan_pipeline} shows the structure of our two-GAN model in which each GAN is a conditional GAN \cite{isola2017image}. The first GAN (segmentation GAN) segments foreground (\eg grassland in soccer games) areas from the input RGB image and outputs a mask image. The second GAN (detection GAN) detects field markings from the foreground image. Each GAN has a discriminator network. The segmentation discriminator predicts whether the mask image is real or fake. The detection discriminator predicts whether the edge image is real or fake. 

The two-GAN model is trained from scratch. The two GANs are first trained independently and then are merged via joint training. In training, we found the segment boundary (grassland vs. non-grassland) has considerable influence on the detection. When a binary segmentation boundary is used, the detection GAN tends to memorize the boundary and that will cause artifacts in testing. We solve this problem by using a soft (alpha-blending) boundary. The hard foreground is 1 and the hard background is 0, with linearly interpolated values between them. We randomly set the width of the alpha-blending band in the range of $[30, 50]$ pixels to prevent the detection GAN from memorizing the bandwidth. 

\subsection{Camera Pose Optimization}
Given a detected edge image, we first extract its feature using the siamese network from Section \ref{sub_sec:deep_feature}. Then we retrieve the initial camera pose from the feature-pose database. After that, we optimize the camera pose by applying the Lucas-Kanade algorithm \cite{baker2004lucas} on two distance images that are from the testing edge image and the retrieved edge image. 

We use a truncated distance (\ie $L2$ norm) images \cite{felzenszwalb2012distance} to overcome the narrow range of gradients in edge images \cite{carr2012point,rematas2018soccer}. With the distance image, we use the Lucas-Kanade algorithm \cite{baker2004lucas} to estimate the homography matrix from the nearest neighbor image to the testing image and use the chain rule to get the refined camera pose.

\section{Evaluation and Experiments}
{\bf World Cup dataset}
This dataset was collected by Homayounfar \etal \cite{homayounfar2017sports} from World Cup 2014. The dataset has 10 games of 209 images for training and 186 images from 10 other games for testing. The images consist of different views of the field with different grass textures, lighting patterns and heavy shadows. We pre-process the training set to obtain following camera configurations: camera location distribution $\mathcal{N}(\mu, \sigma^2)$ ($\mu\approx[52, -45, 17]^\T$ and $\sigma \approx \pm[2, 9, 3]^\T$ meters); pan, tilt and focal length ranges ($[-35^{\circ},35^{\circ}]$, $[-15^{\circ}, -5^{\circ}]$ and $[1000, 6000]$ pixels, respectively). We found these camera parameters are typical settings for many cameras in soccer games.

{\bf Error metric}
We use the intersection over union (IoU) score to measure the calibration accuracy of different methods. The IoU is calculated by warping the projected model to the top view by the ground truth camera and the estimated camera. Homayounfar \etal \cite{homayounfar2017sports} measures the IoU on the whole area of the model, while Sharma \etal \cite{sharma2018automated} measures the IoU only on the area that is visible in the image. We denote these two metrics by IoU$_{whole}$ and IoU$_{part}$, respectively. For a fair comparison, we report both metrics for our method.

\subsection{Synthetic Data Experiments}
\begin{figure}[t]
	\centering
	\includegraphics[width=1.0\linewidth]{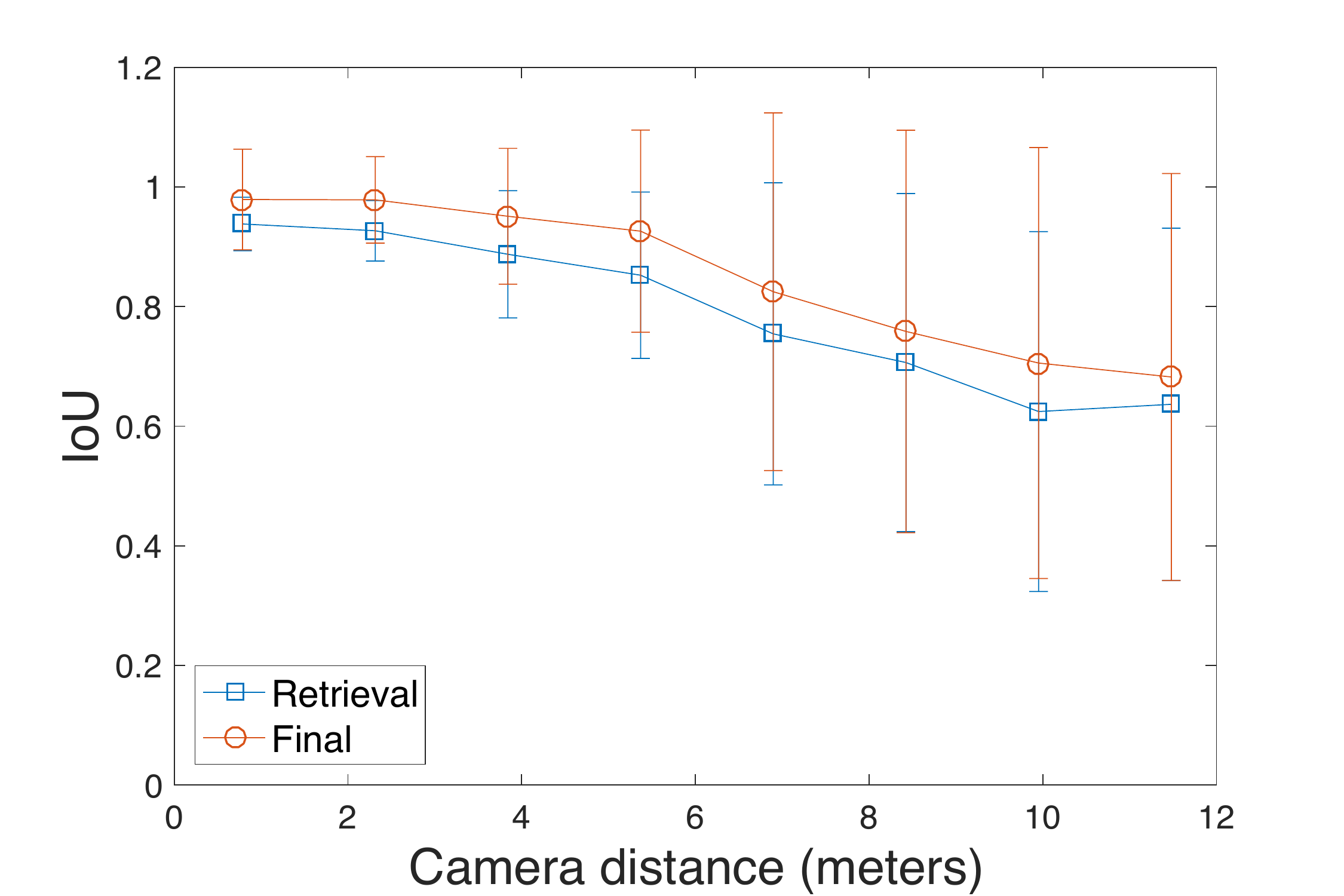}
	\caption{Calibration accuracy (IoU$_{part}$) as a function of quantized camera distance. The blue line shows the retrieval results. The red line shows the refined results. Error bars show standard deviation. Best viewed in color.}    
	\label{fig:iou_vs_dist}
    \vspace{-0.15in}
\end{figure}

We conducted experiments on synthetic data to test the robustness of our method. We fix the training camera location and we vary the displacement between the testing camera and the training camera. This experiment measures the robustness of our method to the displacement of the camera.

First, we generate a training set in which the cameras are located at the same mean location $\mu$ as the World Cup dataset. Then, we generate a testing dataset whose camera locations are $\mu + \mathbf{d}$ in which $\mathbf{d}$ is a random displacement. We randomly generate other parameters (\eg pan, tilt and focal length ranges) as those on the World Cup dataset. For example, the focal length is uniformly sampled from the range of $[1000,6000]$. The training dataset size is 10,000 and the testing dataset size 1,000. The image resolution is $1280 \times 720$. Then, we estimate the camera poses of the testing set using our method. We report the calibration accuracy as a function of camera distances (L2 norm of camera displacement $\mathbf{d}$).

Figure \ref{fig:iou_vs_dist} shows the results. In the figure, the horizontal axis is the quantized camera distance and the vertical axis is the calibration accuracy (mean and standard deviation of IoU$_{part}$). The figure shows that our method achieves high accuracy (IoU$_{part}$ $\geq$ 92\%) when the camera displacement is within 5 meters. We believe this is useful in practice because the camera location generally changes its location in a small range for different games. When the camera location is far from the ground truth camera (\eg about 10 meters), the calibration accuracy gradually drops to about 70\%. 

\subsection{Real Data Experiments}
We evaluate our method on the World Cup dataset and compare with state-of-the-art methods \cite{homayounfar2017sports} and \cite{sharma2018automated}.

\paragraph{Training and testing process}
In training, we sample 100,000 camera poses using our camera pose engine. The camera centers are sampled from the Gaussian distribution $\mathcal{N}(\mu, \sigma^2)$. The pan, tilt and focal length values are sampled from the uniform distributions of  $[-35^{\circ},35^{\circ}]$, $[-15^{\circ}, -5^{\circ}]$ and $[1000, 6000]$ pixels, respectively. The tilt of camera base $\phi^\prime$ is fixed ($-90^{\circ}$) and the roll angle is a random value from $[-0.1^{\circ}, 0.1^{\circ}]$.

We generate edges images ($1280\times720$ resolution) for sampled camera poses. Then, the edge images are resized to $320\times180$ and are used to train a siamese network. The output of the siamese network is a 16-dimension deep feature. The deep features and the camera poses are stored in a feature-pose database.

In testing, we first detect field markings using the two-GAN model on the resized images ($256\times 256$). Then, we query an initial camera pose from the feature-pose database using the deep feature of the detected edge image. After that, we refine the camera pose by the distance image and the LK algorithm.   

\begin{table}[h]
 \centering
 \scalebox{1.0}{
  \begin{tabular}{| l | c| c | c | c |}   
    \hline   
    & \multicolumn{2}{c|}{IoU$_{whole}$} & \multicolumn{2}{c|}{IoU$_{part}$} \\ \cline{2-5}
    Method & Mean & Median & Mean & Median \\ \hline
    
    DSM \cite{homayounfar2017sports}      & 83    & -- & -- & --\\
    Dict. + HOG \cite{sharma2018automated} & -- & -- & 91.4 & 92.7 \\ \hline
    Ours            & \textbf{89.4} & \textbf{93.8} & \textbf{94.5}    & \textbf{96.1}  \\ \hline
  \end{tabular}
  }
  \vspace{1mm}
  \caption{Comparison on the World Cup dataset \cite{homayounfar2017sports}. The results of DSM and Dict. + HOG are taken from their respective papers. The best performance is highlighted by bold.}
   \label{table:soccer_accuracy}   
\end{table}

\paragraph{Main results}
Table \ref{table:soccer_accuracy} shows the main result on the World Cup dataset. We compare our method with previous methods on mean and median IoU scores. Our method is significantly more accurate (89.4 vs. 83) than the DSM method \cite{homayounfar2017sports} in mean IoU$_{whole}$. Our method is also more accurate (94.5 vs. 91.4 in mean IoU$_{part}$) than \cite{sharma2018automated}. Comparing with \cite{sharma2018automated}, our method also has a much simple and interpretable camera pose engine.   

\begin{table}
 \centering
  \scalebox{1.0}{
  \begin{tabular}{| l | c |c |}   
    \hline
    Method    &  {Mean IoU$_{part}$ (\%)}  & $\Delta$ \\ \hline
    Ours          &  \textbf{94.5}  &  --  \\
    Ours w/o segmentation          & 90.2  & 4.3   \\ 
    Ours w/o LK warp     & 91.5  & 3.0   \\
    Ours w/o segmentation and LK warp       & 88.9  &  5.6  \\ 
    Ours replace deep feature with HOG     & 94.5  &  0.0  \\
    \hline
  \end{tabular}
  }
  \vspace{1mm}
  \caption{Components analysis on the World Cup dataset. The last column shows the improvements.}
   \label{table:component_soccer}
   \vspace{-0.15in}
\end{table}

Table \ref{table:component_soccer} shows the component analysis of our method. It shows that both the segmentation GAN and the LK warp contribute to the final result. On this dataset, it is interesting to see that the HOG feature is as good as our method in accuracy but is much less compact (1,860 vs. 16). It shows that the HOG feature is a strong baseline on this dataset.

\begin{figure*}[t]
	\centering
	\includegraphics[width=0.96\linewidth]{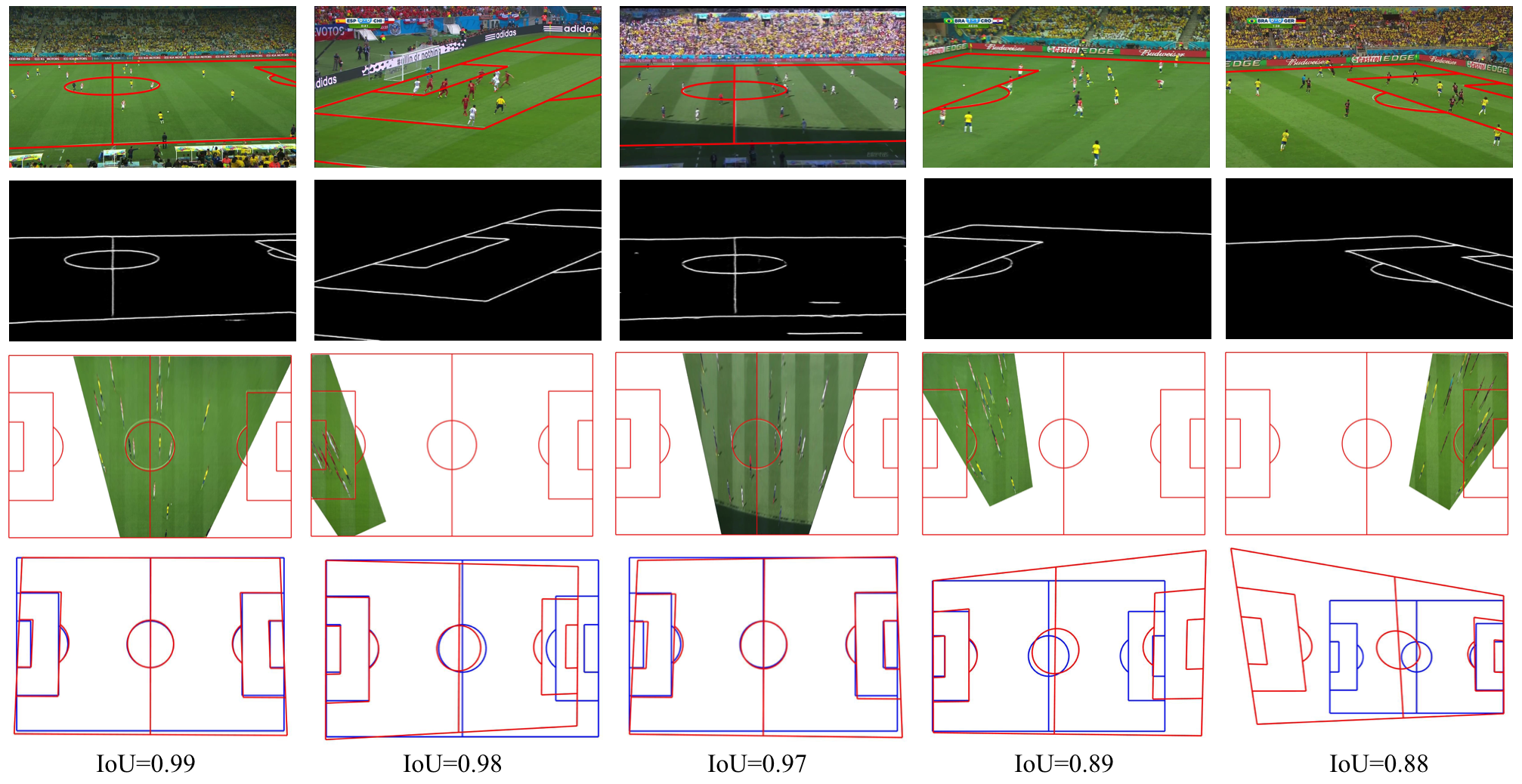}
	\caption[Qualitative results]{Qualitative results. First row: field template overlaid on the image using the estimated camera poses. Second row: detected field markings. Third row: warped image on the template. Fourth row: IoU$_{part}$.}
    \vspace{-0.15in}
	\label{fig:qualitative_results}
\end{figure*}

\paragraph{Qualitative results}
Figure \ref{fig:qualitative_results} shows the qualitative results of our method on the soccer dataset. Our method achieves very high accuracy when the field-of-view is wide (see the first and the third column) or when more field markings are visible (see the second column). When the field markings are not evenly distributed in the image, our method has a lower accuracy (\eg the fifth column in which the bottom touch lines are not visible in the image). We also found that the heavy shadow is the main reason for incorrect field marking detection. However, our method can estimate camera poses in these difficult situations (see the third column). More results are in the supplementary video.

\begin{figure*}[t]
	\centering
	\includegraphics[width=0.96\linewidth]{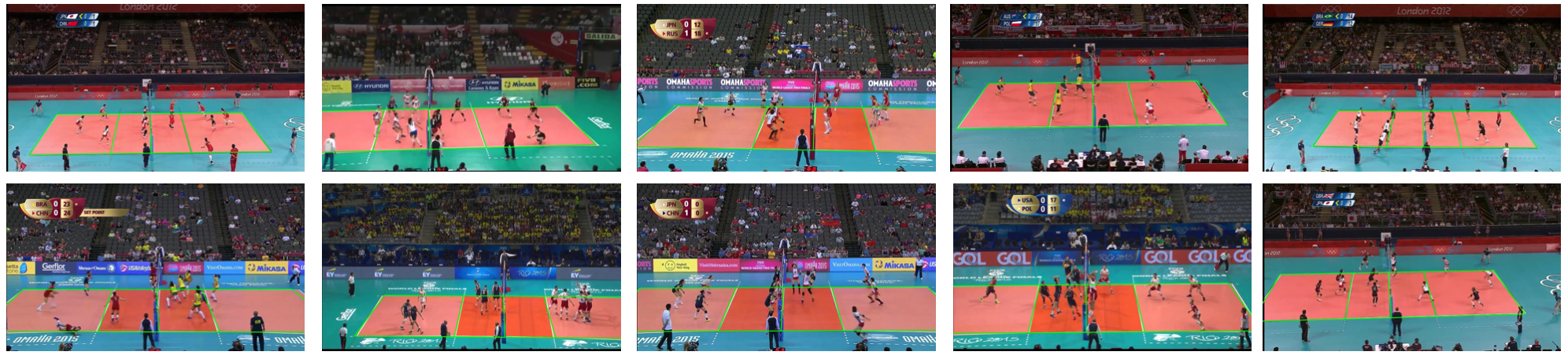}
	\caption{Qualitative results on the volleyball dataset. The green lines are overlaied by estimated camera poses.}
    \vspace{-0.15in}
	\label{fig:volleyball_qualitative}
\end{figure*}

{\bf Volleyball dataset results:} This work focuses on soccer games. However, we also tested our method on a volleyball dataset. The volleyball dataset is collected from the volleyball action recognition dataset of Ibrahim \etal \cite{ibrahim2016hierarchical}. The dataset has 47 games and each game has 10 images. We randomly separated these games into two sets: 24 games for training and 23 games for testing. This data is less challenging than the soccer dataset in terms of lighting conditions. However, it is more challenging in terms of the number of visible field markings. We set the parameters of our method similar to those on the soccer dataset except the camera location, pan, tilt and focal length ranges. Our method achieves very high performance with mean and median IoU$_{part}$ of 97.6\% and 98.8\%, respectively. Figure \ref{fig:volleyball_qualitative} shows qualitative results on the volleyball dataset.

\paragraph{Implementation details}
Our approach (except the two-GAN model) is implemented in Matlab on an Intel 3.0 GHz CPU, 16GB memory Mac system. In the current implementation, the speed is not optimized. In testing, the running time is about 0.5 seconds/frame. We will make the implementation publicly available.

The two-GAN model implementation is based on the pix2pix network \cite{isola2017image} (PyTorch version). In the pix2pix network, both generator and discriminator use modules of the form convolution-BatchNorm-Relu. The generator uses skip connections and follows the shape of a U-Net \cite{ronneberger2015u}. In training, we set the weight of L1 loss $\lambda = 100$. We alternate between one gradient descent step on $D$, then one step on $G$. The losses of the segmentation GAN and the detection GAN have equal weights. We set the batch size to 1 because of limited GPU memory (12GB). The number of epochs is 200 and the learning rate linearly decays from 0.0002. We augment the training set by first resizing the image to $300\times300$ then randomly cropping it to $256\times256$. 

The final result is not sensitive to the parameters of our method. For example, we found 30-50 pixels work well for the distance threshold in the truncated distance image. For the siamese network, the input dimension is $320\times180$ and output dimension is 16. For the similar camera poses, the thresholds of pan, tilt and focal length differences are $1^{\circ}$, $0.5^{\circ}$ and 30 pixels, respectively. 

\paragraph{Discussions}
Although our method achieves the state-of-the-art accuracy on a standard soccer dataset, it has several limitations at the current stage. For example, our method still requires human annotations to train the field marking detection network. One possible improvement is to train the field marking detection network using purely synthetic data, for example, from realistic rendering engines \cite{shafaei2016play}. Moreover, the accuracy of our method relies on the field marking detection result. It fails when there is heavy rain/snow in the playing ground.

\section{Conclusion}
In this work, we proposed an automatic sports camera calibration method by developing a novel camera pose engine. The method is highly automatic and requires very few human interactions. It has achieved the best performance on a standard soccer dataset and very high performance on a volleyball dataset. For future work, we would like to synthesize data from realistic camera pose engines (\eg from sports video games). We also would like to extend our method to other sports such as basketball, ice hockey and US football.








\bibliographystyle{IEEEtran}
\bibliography{icra2019}

\end{document}